\documentclass[conference]{IEEEtran}
\IEEEoverridecommandlockouts

\usepackage[T1]{fontenc} 
\usepackage[utf8]{inputenc} 

\usepackage{cite}
\usepackage{amsmath,amssymb,amsfonts}
\usepackage{algorithmic}
\usepackage{graphicx}
\usepackage{textcomp}
\usepackage{xcolor}

\usepackage{algorithm}

\usepackage{hyperref} 

\usepackage{setspace} 

\usepackage{multirow}

\usepackage{pgf-pie}
\usepackage{caption}
\usepackage{subcaption}

\usepackage{pgfplots}
\usepackage{tikz}
\usepackage{pgfplots}
\pgfplotsset{compat=1.17}

\usepackage{siunitx}
\usepackage{booktabs}

\def\BibTeX{{\rm B\kern-.05em{\sc i\kern-.025em b}\kern-.08em
    T\kern-.1667em\lower.7ex\hbox{E}\kern-.125emX}}

\begin{document}

\title{TRIDIS: A Comprehensive Medieval and Early Modern Corpus for HTR and NER\\


}

\author{\IEEEauthorblockN{1\textsuperscript{st} Sergio Torres Aguilar}
\IEEEauthorblockA{
\textit{University of Luxembourg}\\
Belval, Luxembourg \\
sertor01@ucm.es}

}

\maketitle

\begin{abstract}
This paper introduces TRIDIS (Tria Digita Scribunt), an open-source corpus of medieval and early modern manuscripts. TRIDIS aggregates multiple legacy collections (all published under open licenses) and incorporates large metadata descriptions. While prior publications referenced some portions of this corpus, here we provide a unified overview with a stronger focus on its constitution. We describe (i) the narrative, chronological, and editorial background of each major sub-corpus, (ii) its semi-diplomatic transcription rules (expansion, normalization, punctuation), (iii) a strategy for challenging out-of-domain test splits driven by outlier detection in a joint embedding space, and (iv) preliminary baseline experiments using TrOCR and MiniCPM-Llama3-V 2.5 comparing random and outlier-based test partitions. Overall, TRIDIS is designed to stimulate joint robust Handwritten Text Recognition (HTR) and Named Entity Recognition (NER) research across medieval and early modern textual heritage.
\end{abstract}


\section{Introduction}

The Handwriting Text Recognition (HTR) frameworks are becoming crucial to enlarge access to cultural heritage. Recent large-scale digitization efforts by heritage institutions have enabled the creation of corpora for HTR and downstream tasks like named entity recognition (NER) or morphological analysis. However, many existing datasets suffer from inconsistent annotation standards (e.g., variations in abbreviation expansion, punctuation handling, or allographic normalization), limited scope (e.g., focusing on a single script family, writer or geographical region), or the lack of well-defined and challenging test splits. Furthermore, domain overlaps between training and testing documents are sometimes excessive, preventing realistic out-of-domain performance assessments. These limitations hinder the development and robust evaluation of HTR models, particularly for out-of-domain generalization, as they cannot be used reliably as task benchmarks.

This paper introduces TRIDIS (\emph{Tria Digita Scribunt}), a unified and expanded corpus designed to address these shortcomings. TRIDIS aggregates multiple open-source sub-collections (each with its own DOI) of medieval and early modern manuscripts, alongside new annotations that expand the corpus's chronological and linguistic coverage and correct some errors, into a single, consistently structured resource, organized using a standardized schema and packaged in Apache Parquet format for efficient access and analysis.  While subsets of TRIDIS have been utilized in previous research \cite{aguilar2023handwritten, strobel2024multilingual, aguilar2021named}, this paper provides the first comprehensive description of the complete corpus, its composition, and its unique features.

Crucially, we propose a novel outlier-driven partition strategy. Unlike traditional random splits, which often yield optimistic performance due to domain overlap between training and testing, our approach identifies and isolates challenging examples, characterized by unusual script variations, rare vocabulary, and complex layouts, to define the test set. This provides a more realistic evaluation of HTR model robustness and generalization capabilities beyond in-domain scenarios.

To demonstrate the impact of this approach, we present baseline experimental results using the TrOCR model \cite{li2023trocr} and the MiniCPM-Llama3V 2.5 \cite{yao2024minicpm} as pre-train foundations that illustrate the significant performance gap when evaluating on outlier-driven test splits. This highlights the importance of rigorous evaluation methodologies in HTR research and reveals previously underestimated challenges in HTR training and evaluation.

\section{Related Work}
Historical HTR has progressed during the past decade moving from Hidden Markov Models to deep learning architectures like Convolutional Recurrent Neural Networks (CRNNs)\cite{kiessling2019kraken}, Transformers \cite{li2023trocr}, and more recently, multimodal models integrating visual-language pretraining and generative data augmentation techniques \cite{aguilar2025dual}. Projects such as \emph{Himanis}\cite {stutzmann_dominique_2021_5535306}, \emph{HOME-Alcar}\cite{stutzmann2021home}, \emph{e-NDP}\cite{claustre_2023_7575693}, \emph{Bullinger Digital}\cite{scius2023bullinger} and \emph{CATMuS}\cite{clerice2024catmus} have generated large volumes of line-level ground truth, primarily for Latin and medieval French, High German and Spanish, concentrated on the documentary manuscripts, i.e, manuscripts with a juridical, administrative and managegerial function, crucial sources for history from the 13th century onwards. 

However, challenges remain. Cross-collection data reuse is often complicated by variable transcription guidelines and varying levels of annotation detail \cite{stutzmann2021home}. Furthermore, documentary sources present unique difficulties, including complex layouts, high script variability, ligatures, and extensive abbreviations practices specially before the 14th century. The issue of domain overlap between training and testing data, leading to inflated performance metrics, is also a significant concern as is often demonstrated in works using historical HTR benchmark corpora as Saint-Gall\cite{fischer2011transcription}, Parzival\cite{fischer2009automatic} and Esposalles\cite{romero2013esposalles}.

The corpora on TRIDIS follows the \emph{Semi-diplomatic} transcription paradigm \cite{driscoll }, balancing philological accuracy with readability. This paradigm, common in historical editions, involve expanding all abbreviations, scribal and typographic marks, modernizing punctuation, and normalizing allographic variations (e.g., long-s vs. short-s). In medieval manuscripts, scribes extensively used abbreviations for lexical or grammatical elements\cite{hasenohr1998abreviations}, but modern editors often restore them silently, adopting a modern punctuation scheme to identify clauses, and unify disparate letterforms. This transcription mode has the advantage of facilitating usage of modern NLP pipelines with the content while introducing an interpretative layer that challenge HTR models trained on such data.

Overall, these advances—from CRNNs to Transformers, and further to multimodal systems— reflect a trend toward leveraging comprehensive contextual and cross-modal information. TRIDIS is conceived within this evolving framework, aiming to provide a resource that supports both traditional HTR and downstream NLP tasks, while explicitly addressing the challenges posed by heterogeneous transcription standards. The aggregation of diverse sources, combined with new annotations and a focus on challenging examples, aims to improve the generalizability of HTR models to unseen historical documents.

\subsection{Editorial Transcription Rules in Practice}
The majority of these sub-corpora adopt a \emph{semi-diplomatic} editorial stance:

\begin{itemize}
    \item All Abbreviations are silently expanded (e.g., \emph{d\textsuperscript{ms}} $\rightarrow$ dominus, facim\raisebox{-0.1em}{\includegraphics[height=0.8em]{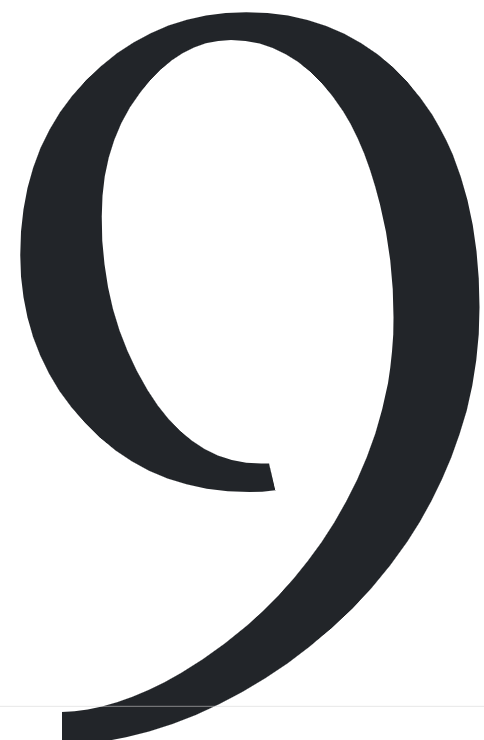}} $\rightarrow$ facimus).
    \item Allograph variants (``long-s'', multiple forms of ``r'') are collapsed into a standard extended Latin letter.
    \item Sentence-level punctuation is typically modern, chosen by editors to reflect textual structure.
    \item Spaces and major and minor pauses have been also modernized with conventional modern punctuation.
    \item Words agglutination, common in book manuscripts, have not been followed.
    \item Special signs (currency, glyphs) remain intact if the editor have determined them as crucial for content.
    \item Notarial marks and conventional signs (\raisebox{-0.1em}{\includegraphics[height=0.8em]{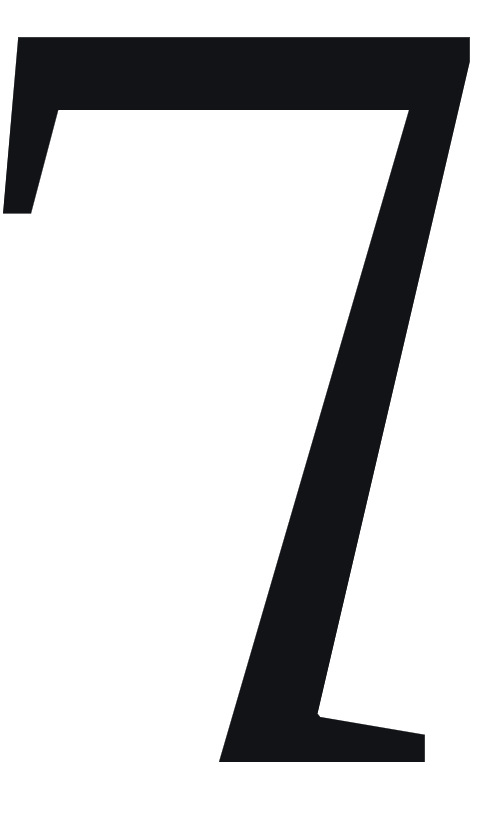}} $\rightarrow$ et ; \raisebox{-0.1em}{\includegraphics[height=0.85em]{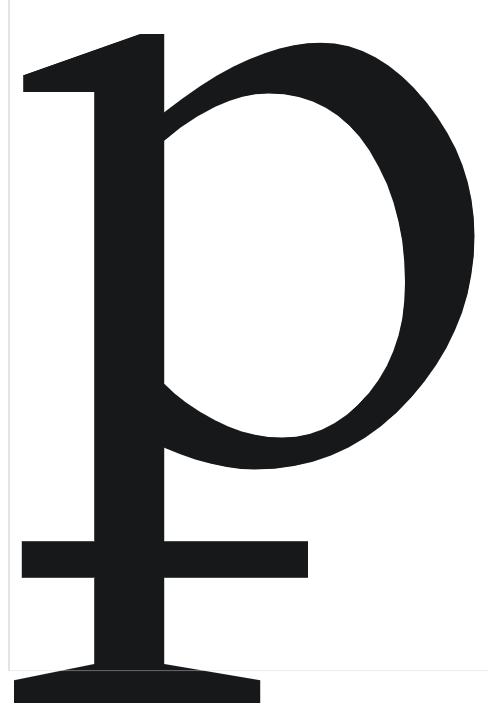}} $\rightarrow$ pro) have been resolved. 
    \item The consonantal ``i'' and ``u'' characters have been transcribed as ``j'' and ``v'' in both French and Latin.
    \item Named entities (names of persons, places and institutions) have been capitalized.
    \item Corrections and canceled words in the manuscript are transcribed enclosed by the sign \$ .
\end{itemize}
While guidelines exist to standardize semi-diplomatic transcriptions, variations inevitably persist across different projects and editorial teams. TRIDIS merges them under a uniform data model that indicates the rules and transcription style in the metadata.

\begin{table*}[htbp]
\centering
\scalebox{1}{ 
\begin{tabular}{|c|c|c|c|S[table-format=3.1]|S[table-format=3.1]|c|}
\hline
\textbf{Sub-corpus} & \textbf{Dates c.} & \textbf{Languages} & \textbf{Script Families} & {\textbf{\#Lines}} & {\textbf{\#Tokens}} & \textbf{Hands}\\
\hline
Alcar-HOME & 12--14th & la, fro & Textualis, Cursiva, Curs.Ant. & 95.6k & 888.2k & >40 \\
\hline
Himanis & 14--15th & la, fro & Cursiva, Hybrida & 22.0k & 417.2k & >10 \\
\hline
e-NDP & 14--15th & la, fro & Cursiva & 33.5k & 217.3k & $\sim$18 \\
\hline
CODEA & 11--16th & la, spa & Cursiva, Humanistic & 5.5k & 57.7k & $\sim$22 \\
\hline
Bullinger & 15--16th & la, gmh & Cursiva, Humanistic & 16.5k & 127.6k & >10 \\
\hline
Koenigsfelden & 14--16th & la, deu & Textualis, Cursiva & 3.0k & 65.8k & >20 \\
\hline
MLH & 12--16th & la, fro, gmh & Textualis, Cursiva, Pr.Gothic & 11.0k & 131.1k & >15 \\
\hline
\multicolumn{6}{l}{\footnotesize * Additional tiny corpora are omitted for brevity. ``\#Hands'' is approximate.}\\
\end{tabular}}
\caption{Selected Sub-Corpora in TRIDIS: Chronology, Languages, Script Families, Lines, Tokens and Hands}
\label{tab:composition}
\end{table*}

\section{Constitution of the TRIDIS Corpus}
\subsection{Overview of the Major Sub-Corpora}

TRIDIS sub-corpora were selected for: 

\begin{enumerate}
    \item Open Licenses: Ensuring broad accessibility (CC BY, CC BY-SA).
    \item Representativeness: Covering major Western European scripts (Cursiva, Textualis) and languages (Latin, Old French, Middle High German, Old Spanish) from the 12th-17th centuries, with smaller cross-periods subsets for added diversity.
    \item Challenging Document Types: The focus is on documentary sources (charters, registers, letters), which feature complex layouts and significant handwriting variation.
    \item Complementarity: Aiming for broad chronological, geographical, and typological coverage. Sub-corpora were chosen to enrich overall diversity.
\end{enumerate}

While biases exist, TRIDIS provides a balanced and challenging benchmark, especially for documentary sources since the late 13th-century. Corpora composing the corpus are described as follows:

\textbf{Alcar-HOME (12--15th c.)}: Cartularies featuring Latin and Old French. These volumes date to a transitional phase in script usage, encompassing Textualis for earlier pages and evolving into Cursiva forms in the later 13th century. Most acts revolve around property transfers, wills, privileges and diplomatics letters. Is the only corpus presenting HTR and NER aligned annotations. \cite{stutzmann2021home}.

\textbf{Himanis (14--15th c.)}: Royal registers from the French Royal Chancery Written in medieval Latin and French. They employ a notarial Cursiva style interspersed with Hybrida. These registers expand on 75k pages, containing thousands of annotated lines capturing official communications (letters of remission, mandates, accounts, etc.) \cite{stutzmann_dominique_2021_5535306}.

\textbf{e-NDP (14--15th c.)}: 24 Registers from the cathedral chapter of Notre-Dame de Paris, primarily in Latin with intermittent French from the mid-14th to the early 15th century, in cursive script. They include official deliberations on liturgical, financial, community and management matters of a main French actor in the Middle Times. \cite{claustre_2023_7575693}.

\textbf{CODEA (11--16th c.)}: Spanish documentary sources bridging Medieval Spanish and Latin usage, often revealing the shift toward humanistic and process cursive in the 15th century. Documents collected range from notarial acts to regal decrees and even witchcrafts. \cite{borja2012desarrollo}.

\textbf{Bullinger (15--16th c.)}: Reformation-era private correspondence in Latin and Early New High German. Script ranges from Cursiva to a transitional or more modern humanistic style. We choose a random sub-set of 10\% of the full corpora for each language. \cite{scius2023bullinger}.

\textbf{Köenigsfelden (14--16th c.)}: Charters from the Swiss Abbey of Köenigsfelden. Latin and Middle High German are present. Script is dominantly Textualis for earlier items and more cursive forms for later expansions \cite{halter_pernet_2021_5179361}.

\textbf{VOC (17--18th c.)}: Macro ground-truth collection of 6000 pages gathered by the the The National Archives of the Netherlands. The documents come from the 17th and 18th century archives from the Dutch East-India Company (VOC). Only a tiny sub-corpora from 17th documents were integrated. \cite{https://doi.org/10.5281/zenodo.6414086}.

\textbf{Monumenta Luxemburgensia Historica (12th-16th)} : Charters and registers coming from the Luxembourg Duchy in Latin, High Middle German and Medieval French. This corpus spans different typologies: cartularies, feudal books and early state registers.

Some other tiny sub-corpora are included, such as manuscript book pages, test pages from local archives, or 17th-century transitional items to reinforce the diversity and gray holes periods. Altogether, TRIDIS account for almost 200k lines and 2M tokens distributed across different institutions, centuries and document typologies.

\subsection{Table of Composition and Hand Counts}
To illustrate scale and variation, Table~\ref{tab:composition} presents an illustrative summary of the main sub-collections included in TRIDIS, including stats about Languages, chronologies, script families and the number of distinct scribal hands. (Scribal hand detection is approximate, based on paleographic or editorial notes from each original sub-corpus.)

\begin{figure}[htbp]
  \begin{subfigure}[t]{0.43\textwidth}
    \raggedleft
    \begin{tikzpicture}
      \begin{axis}[
          xbar,
          width=6.6cm,
          height=3.2cm,
          xlabel={(a) Languages \%},
          symbolic y coords={Latin, French, German \& Dutch, Spanish},
          ytick=data,
          nodes near coords,
          nodes near coords style={anchor=west, xshift=1mm},
          xmin=0, xmax=60,
          bar width=0.3cm,
          enlarge y limits=0.25,
          axis y line*=left,
        ]
        \addplot coordinates {(57.5,Latin) (29,French) (9,German \& Dutch) (4.5,Spanish) };
      \end{axis}
    \end{tikzpicture}
  \end{subfigure}
        
\hfill
\begin{subfigure}[t]{0.4\textwidth}
  \raggedright
  \begin{tikzpicture}
    \begin{axis}[
        xbar,
        width=6.7cm,
        height=4.5cm,
        xlabel={(b) Century \%},
        symbolic y coords={$13^{\text{th}}$, $15^{\text{th}}$, $14^{\text{th}}$, $16^{\text{th}}$, $12^{\text{th}}$, $17^{\text{th}}$, Others},
        ytick=data,
        nodes near coords,
        nodes near coords style={anchor=west, xshift=1mm},
        xmin=0, xmax=35,
        bar width=0.3cm,
        enlarge y limits=0.15,
        axis y line*=left,
      ]
      \addplot coordinates {(31.5,$13^{\text{th}}$) (22.5,$15^{\text{th}}$) (20.5,$14^{\text{th}}$) (9,$12^{\text{th}}$) (4.5,$17^{\text{th}}$) (11.5,$16^{\text{th}}$) (0.5,Others)};
    \end{axis}
  \end{tikzpicture}
\end{subfigure}
  
  \vspace{1ex}
  
  \begin{subfigure}[t]{0.4\textwidth}
    \raggedleft
    \begin{tikzpicture}
      \begin{axis}[
          xbar,
          width=6.7cm,
          height=4cm,
          xlabel={(c) Scripts \%},
          symbolic y coords={Cursiva, Textualis, Cursiva Antiqua, Semi-Hybrida, Praegotica, Others},
          ytick=data,
          nodes near coords,
          nodes near coords style={anchor=west, xshift=1mm},
          xmin=0, xmax=60,
          bar width=0.3cm,
          enlarge y limits=0.2,
          axis y line*=left,
        ]
        \addplot coordinates {(59,Cursiva) (27.5,Textualis) (6,Cursiva Antiqua) (4.5,Semi-Hybrida) (2.5,Praegotica) (0.5,Others)};
      \end{axis}
    \end{tikzpicture}
  \end{subfigure}
  
  \caption{Percentual distribution of Languages, chronologies and Script Families in TRIDIS}
  \label{fig:bar-charts}
\end{figure}
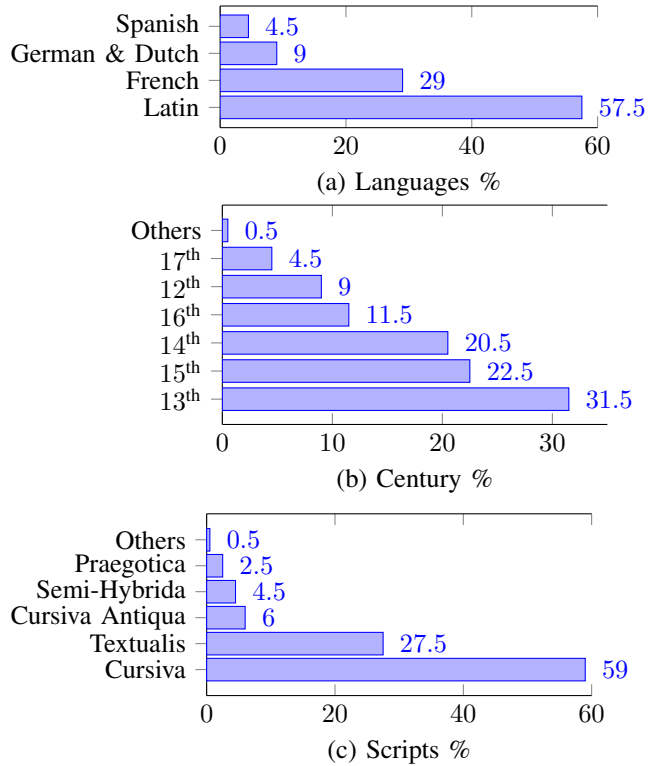

\subsection{Source Data Overview}
Each sub-corpus was originally released via open-access platforms, typically with its own DOI. They cover manuscripts dated from the late 10th to early 17th century, in multiple script families:

All textual lines are stored in a single Parquet file. Each record retains data about:
\begin{itemize}
    \item \texttt{Image}: Binary version of the line image in RGB encoding.
    \item \texttt{Text} : Text of the graphical line in UTF-8
    \item \texttt{Manuscript\_id}: short name or ID referencing the manuscript name or sub-corpus.
    \item \texttt{Language}: Dominant Language in the line or page.
    \item \texttt{Century}: Century of the manuscript.
    \item \texttt{script\_family}: Dominant written family in the line or page.
    \item \texttt{NER\_annotation}: BIO annotation of named entities when they are available or null value.
\end{itemize}
In some rare cases lines remain bilingual or partially code-switched, in other cases they are only composed of a name, a number or date, especially in certain documentary sources. Furthermore is not uncommon that a line couldn't be certainly classified in a script family, on these cases each line is tagged with the dominant language and script in the page for approximate classification purposes.

The dataset splits contains:
\begin{enumerate}
    \item Train set : \hspace{4mm}177 660 lines
    \item Val set : \hspace{10.7mm}9 829 lines
    \item Test set : \hspace{9.6mm}9 827 lines
\end{enumerate}

\subsection{Availability and Licensing}
TRIDIS is publicly deposited in HuggingFace, under the name \texttt{\url{magistermilitum/Tridis}}. The code repositories also provide instructions on how to convert or combine the line-level data with digital facsimile images. Each sub-corpus retains its original license---commonly CC BY or CC BY-SA---and the aggregator repository clarifies the relevant references and usage terms.

\section{Outlier-Based Splits and Baseline Experiments}
\subsection{Outliers Splitting Strategy}

To recreate this scenario more faithfully, we compute joint embeddings for each line by concatenating the mean-pooled outputs from its image and text encoders (see Algorithm~\ref{algo:outlier}). These embeddings capture both the visual (e.g., layout, handwriting) and linguistic features. We then define a centroid as the element-wise median of all embeddings, representing the corpus’s central tendency. Lines with the largest Euclidean distances from this centroid (the top 5\%) are labeled as "outliers" and form the test partition. To ensure diversity, we stratify these outliers by distance, selecting a representative sample from different levels of "atypicality". This mimics the unpredictable nature of new historical documents.

\begin{figure}[htbp]
    \centering
   \scalebox{1.155}{ 
    \includegraphics[width=0.85\linewidth]{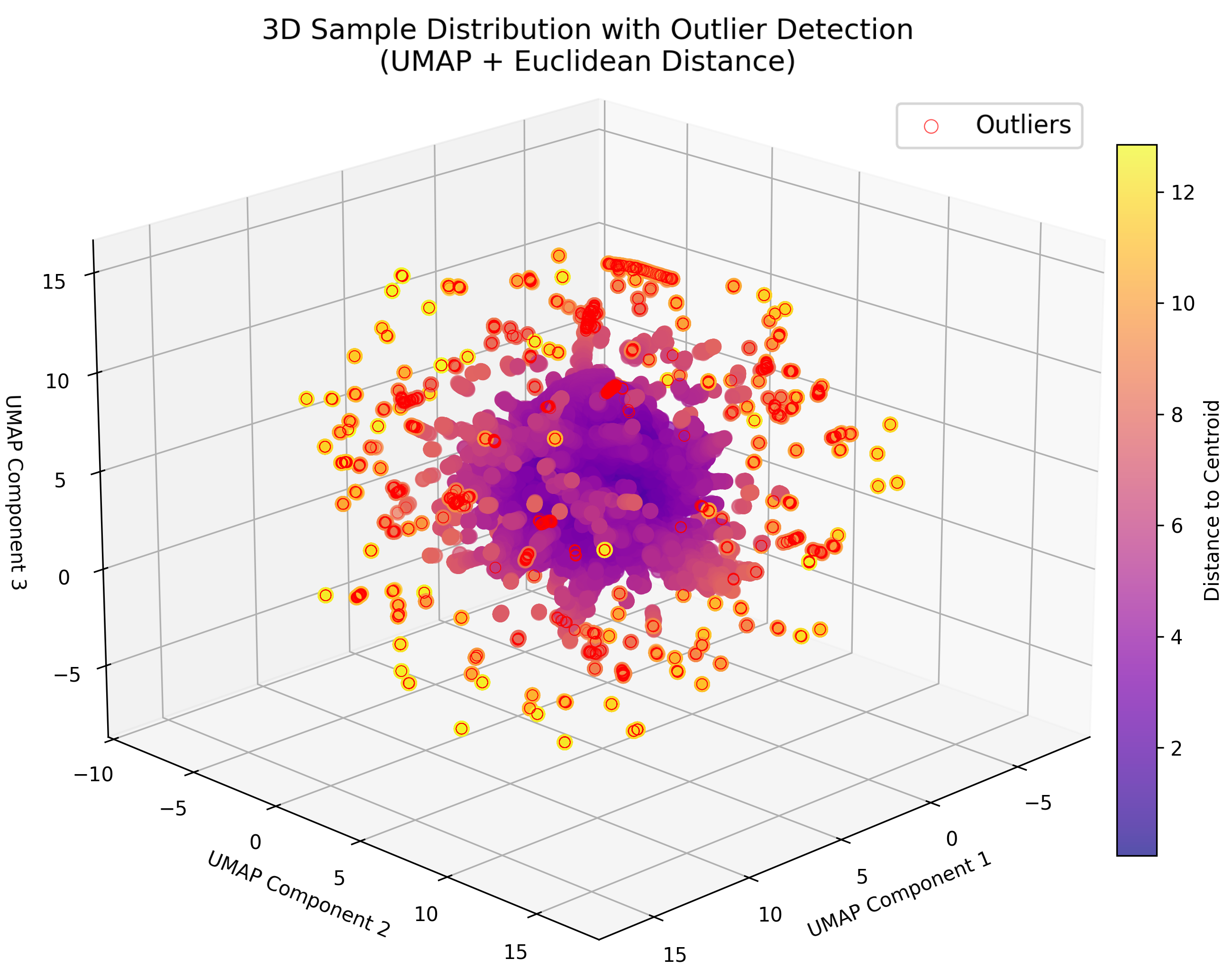}}
    \caption{3D UMAP distribution Points in red circles are outliers assembled for the test set. This group exhibiting high density of challenge features are typically more than 9 units away from the centroid.}
    \label{fig:3Doutliers}
\end{figure}

In short, samples far from the centroid tend to have a higher density of rare or challenging characteristics (see Taxonomy section), making this outlier-based split a more rigorous measure of HTR robustness than random splits. For clarity, Figure~\ref{fig:3Doutliers} demonstrates the distribution of outliers in a 3D UMAP plot.

\subsection{Baselines: TrOCR vs.\ MiniCPM-Llama3-V 2.5}
As an illustration, we trained two baseline HTR models:
\begin{itemize}
    \item \textbf{TrOCR (Large)} \cite{li2023trocr}: A Transformer-based encoder-decoder approach designed for OCR/HTR tasks using Vit + RoBERta. (558M parameters). Trained during 15 epochs (Linear, 1e-4)
    \item \textbf{MiniCPM-Llama3-V 2.5} \cite{yao2024minicpm}: A smaller multilingual Vision+Language architecture, also adapted for HTR line recognition. (8B parameters). Trained during 4 epochs (Cosine, 5e-5)
\end{itemize}

\begin{table}[htbp]
\centering
\scalebox{0.85}{ 
\begin{tabular}{l|c|c|c}
\hline
\textbf{System / Split} & \textbf{CER (\%)} & \textbf{WER (\%)} & \textbf{BERT Sc}\\
\hline
TrOCR (Random) &  9.1 & 21.3 & 0.94\\
TrOCR (Outlier) & 11.3 & 24.9 & 0.91\\
\hline
MiniCPM2.5 (Random) & 10.2 & 24.2 & 0.91\\
MiniCPM2.5 (Outlier) & 12.6 & 28.0 & 0.89\\
\hline
\end{tabular}}
\caption{Placeholder Baseline Results on TRIDIS (Random vs.\ Outlier-based Test).}
\label{tab:baseline}
\end{table}

Both models were fine-tuned on 90\% of the corpus, with 5\% reserved for validation and the remaining 5\% identified as outliers forming the test set. Table~\ref{tab:baseline} shows placeholder results comparing a conventional random-split test to our more challenging outlier-based split. Consistent with literature findings \cite{clerice2024catmus}, the outlier test partition substantially increases the difficulty, underscoring the domain drift characteristic of newly encountered medieval manuscripts.

Under typical in-domain conditions for semi-diplomatic transcriptions, current state-of-the-art HTR models usually achieve a Character Error Rate (CER) near or immediately below 10\% (9\% in our case), which maintains acceptable machine and human-readability (94\% of semantic similarity according to BERT score). However, as demonstrated in related work \cite{aguilar2025dual, alkendi2024advancements}, CER can rise to 15--25\% for out-of-domain manuscripts—even if they share approximate chronology or script families. Notably, the CER metric in our experiments counts errors associated with spaces and punctuation, which are one of the main sources of confusion in automatic transcriptions.

\begin{figure*}[htbp]
    \centering
   \scalebox{0.85}{ 
    \includegraphics[width=0.85\linewidth]{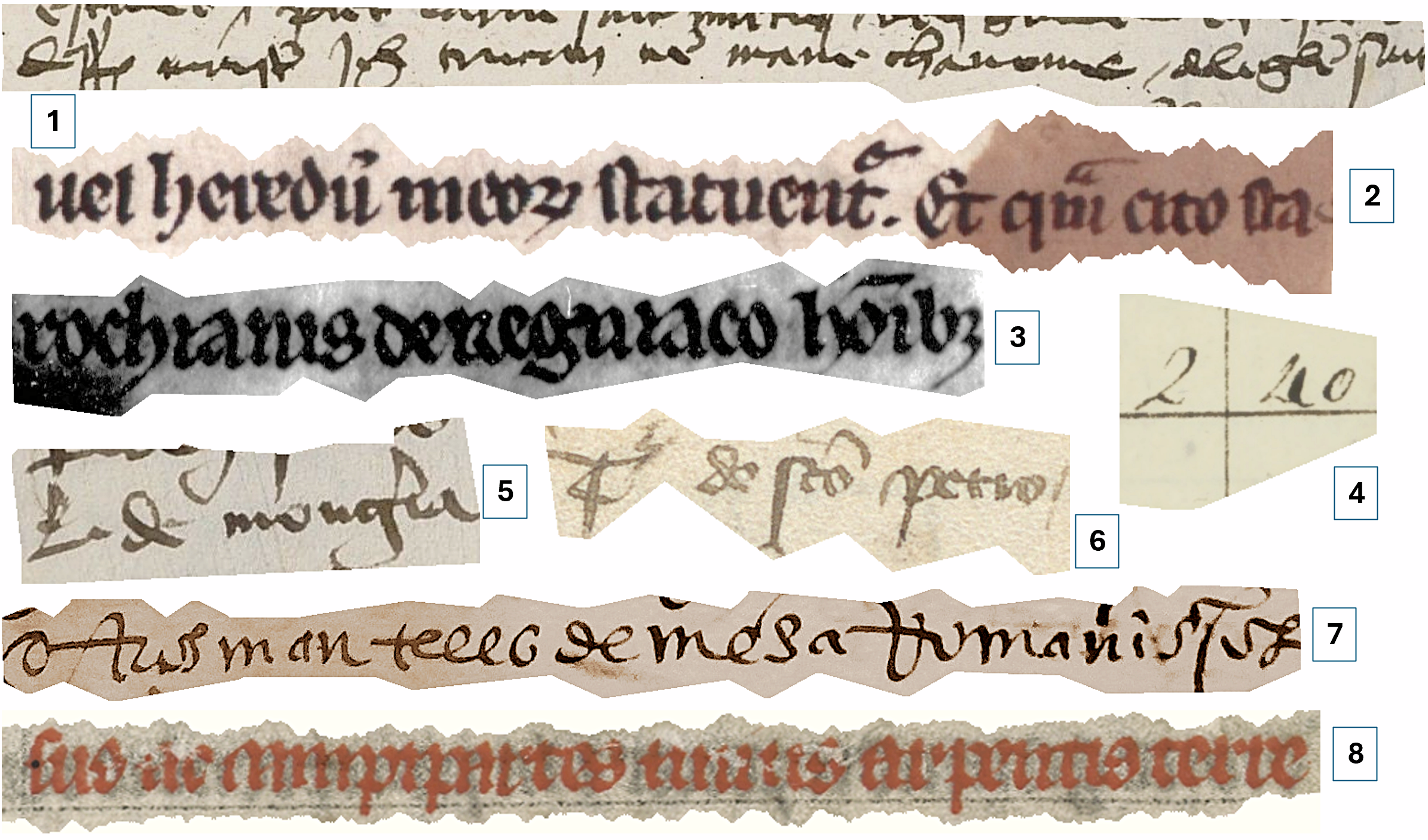}}
    \caption{Examples of outliers lines from the TRIDIS test set: 1. defunct maistre Jehan Trucan ne mane chanoine / de l'eglise saint. || 2. vel heredum meorum statuentur, et, quam cito sta- || 3. rochianis de Regniaco hominibus || 4. 2 - 40  || 5. L. de Mongeria || 6. T. de Sancto Petro || 7. otros manteles de mesa Romaniscos || 8. sus ac campipartes unius arpentis terre}
    \label{fig:tridis_outliers}
\end{figure*}

\begin{algorithm}[H]
\caption{Outlier Detection and Stratified Splitting}
\label{algo:outlier}
\begin{algorithmic}[1]
\STATE \textbf{Input:} Dataset $D=\{(I_i,T_i)\}_{i=1}^N$, outlier ratio $p = 0.04$, validation ratio $r_{val} = 0.05$.
\FOR{$i=1$ \TO $N$}
    \STATE $e_i \leftarrow [\,\text{VisionEncoder}(I_i);\,\text{TextEncoder}(T_i)]$
\ENDFOR
\STATE Standardize $\{e_i\}$; centroid $C \leftarrow \text{median}(\{e_i\})$.
\FOR{$i=1$ \TO $N$}
    \STATE $d_i \leftarrow \|e_i - C\|_2$
\ENDFOR
\STATE Sort distances $d$ descending; $T \leftarrow d_{\lceil p \cdot N \rceil}$
\STATE $S \leftarrow \{\}$; Divide $d$ into $B$ bins.
\FOR{each bin $b$ in $B$}
  \STATE $S \leftarrow S \cup \text{RandomSubset}(\{D_i \mid d_i \in b, d_i \geq T\}, \lfloor p \cdot |b| \rfloor)$
\ENDFOR
\STATE $D_{test} \leftarrow S$
\STATE Split $D \setminus D_{test}$ into $D_{train}$ and $D_{val}$ (ratio $r_{val}$).
\STATE \textbf{Output:} $D_{train}$, $D_{val}$, $D_{test}$.
\end{algorithmic}
\end{algorithm}

\subsection{Outliers Taxonomy}

In our analysis of the outlier test set, several recurrent issues have been identified that contribute to increased transcription difficulties. It is worth noting that these challenges are also observable in the other two corpus subsets; however, their concentration is significantly higher in the outlier test set. We briefly show main categories in Fig.~\ref{fig:tridis_outliers} and describe they as follows:

\begin{itemize}
    \item \textbf{Names:} Lines that contain primarily or exclusively proper names (e.g., examples 5 and 6) are especially challenging because they fall outside the standard vocabulary and typically exhibit strong abbreviation practices. For instance, "T. de Sancto Pedro" may correspond to various names (such as Thomas, Teobald, or Theodulf).
    
    \item \textbf{Long Lines:} Lines that are unusually long (exceeding 20 words, compared to a mean of 7 words per line) or that display variable geometry—such as changes in direction or infiltrations from adjacent lines (e.g., example 1)—pose significant difficulties for transcription.
    
    \item \textbf{Physical or Scanning Defects:} Lines affected by physical issues, including transparencies, weak or nearly faded ink, stains, smudges, distorsions or canceled content (as seen in examples 2, 3, and 8), impede accurate recognition.
    
    \item \textbf{Underrepresented Script Families:} Lines belonging to less frequently observed script families (e.g., Praegotica, Humanistica, Carolingia) are particularly problematic due to the limited number of examples available for model training (as in examples 8 and 7).
    
    \item \textbf{Extremely Challenging Cases:} The most common difficulty arises in lines from the 16th notarial hands, which often exhibit a particularly intricate procedural handwriting style (e.g., example 1 et 7). Additional challenges are found in lines associated with marginalia, page borders, indices, etc. (e.g., example 4).
\end{itemize}

\section{Discussion}
 
TRIDIS is expressly designed for flexible usage and fair evaluation. It not only unifies thousands of lines from medieval and early modern manuscripts—spanning multiple centuries, languages, and script families—but also encodes rich metadata (e.g., named entities, manuscript chronologies, dominant script, and language). This coverage enables cross-linguistic analyses, style-shift investigations, and the study of paleographic nuances across regions, while the out-of-domain lines illustrate how local scribal traditions can evolve unpredictably. Moreover, many downstream tasks require robust multimodal representations; thus, a corpus that integrates image and text embeddings—enriched by extensive metadata—becomes instrumental for developing advanced HTR modules involving script classification, named entity recognition (NER), and deeper document understanding. 

Our experimental results confirm that an outlier-based test set poses greater challenges for HTR models than a random partition. This gap highlights not only the models’ sensitivity to rare or atypical factors (e.g., unusual abbreviations or physically degraded pages), but also demonstrates a systematic way to expose current HTR limitations, evidenced by higher CER ($>$2 points) and WER ($>$4 points) scores than in random splits. For heritage institutions digitizing new manuscripts, the new data often lies outside the immediate distribution of the training set. Hence, an outliers test set offers a more realistic measure of out-of-domain performance and encourages the creation of more adaptable solutions. Moreover, analyzing these outliers provides insights into scribal diversity, suggesting targeted augmentation strategies—e.g., for underrepresented scripts—or specialized normalization approaches for complex abbreviations.

\section{Conclusion}

We have presented TRIDIS, a comprehensive corpus of medieval and early modern manuscripts aggregated from multiple open repositories. By embedding rich metadata on languages, centuries, writing families, and named entities, TRIDIS offers applications that extend beyond conventional HTR tasks.

Further, our outlier-based partition approach yields a more demanding test scenario that better reflects the challenges of manuscripts exhibiting rare or unconventional features, revealing domain adaptation issues that remain hidden under random splits.

Future work involves including additional cross-language and under-represented scripts collections, refining aligned transcriptions, and expanding metadata categories. Less common scripts (such as Carolingian minuscule) or ancient language states are often absent from large training sets, leading to poor performance when encountered in new documents. Ultimately, TRIDIS serves as a foundational resource and baseline for novel methodologies, promoting cross-domain transfer in the computational study of heritage documents.

\section{Repositories}
The models supporting this study are available under open source licenses: 

\vspace{1ex}
Baseline models:

TrOCR : 

{\footnotesize \url{https://huggingface.co/magistermilitum/tridis_HTR}}

MiniCPM: 

{\footnotesize \url{https://huggingface.co/magistermilitum/Tridis_HTR_MiniCPM}}

\vspace{1ex}

Corpora: 

{\footnotesize \url{https://huggingface.co/datasets/magistermilitum/Tridis}}

\section*{Acknowledgments}
The authors would like to thank the original institutions and researchers who released each sub-corpus under open licenses, making the creation of TRIDIS possible.

\bibliographystyle{IEEEtran}
\bibliography{main}  

\begin{thebibliography}{10}
\providecommand{\url}[1]{#1}
\csname url@samestyle\endcsname
\providecommand{\newblock}{\relax}
\providecommand{\bibinfo}[2]{#2}
\providecommand{\BIBentrySTDinterwordspacing}{\spaceskip=0pt\relax}
\providecommand{\BIBentryALTinterwordstretchfactor}{4}
\providecommand{\BIBentryALTinterwordspacing}{\spaceskip=\fontdimen2\font plus
\BIBentryALTinterwordstretchfactor\fontdimen3\font minus \fontdimen4\font\relax}
\providecommand{\BIBforeignlanguage}[2]{{%
\expandafter\ifx\csname l@#1\endcsname\relax
\typeout{** WARNING: IEEEtran.bst: No hyphenation pattern has been}%
\typeout{** loaded for the language `#1'. Using the pattern for}%
\typeout{** the default language instead.}%
\else
\language=\csname l@#1\endcsname
\fi
#2}}
\providecommand{\BIBdecl}{\relax}
\BIBdecl

\bibitem{aguilar2023handwritten}
S.~Torres~Aguilar and V.~Jolivet, ``Handwritten text recognition for documentary medieval manuscripts,'' \emph{Journal of Data Mining and Digital Humanities}, 2023.

\bibitem{strobel2024multilingual}
P.~B. Str{\"o}bel, L.~Fischer, R.~M{\"u}ller, P.~Scheurer, B.~Schroffenegger, B.~Suter, and M.~Volk, ``Multilingual workflows in bullinger digital: Data curation for latin and early new high german,'' \emph{Journal of Open Humanities Data}, vol.~10, no.~12, p.~12, 2024.

\bibitem{aguilar2021named}
S.~T. Aguilar and D.~Stutzmann, ``Named entity recognition for french medieval charters,'' in \emph{Proceedings of the Workshop on Natural Language Processing for Digital Humanities}, 2021, pp. 37--46.

\bibitem{li2023trocr}
M.~Li, T.~Lv, J.~Chen, L.~Cui, Y.~Lu, D.~Florencio, C.~Zhang, Z.~Li, and F.~Wei, ``Trocr: Transformer-based optical character recognition with pre-trained models,'' in \emph{Proceedings of the AAAI Conference on Artificial Intelligence}, vol.~37, no.~11, 2023, pp. 13\,094--13\,102.

\bibitem{yao2024minicpm}
Y.~Yao, T.~Yu, A.~Zhang, C.~Wang, J.~Cui, H.~Zhu, T.~Cai, H.~Li, W.~Zhao, Z.~He \emph{et~al.}, ``Minicpm-v: A gpt-4v level mllm on your phone,'' \emph{arXiv preprint arXiv:2408.01800}, 2024.

\bibitem{kiessling2019kraken}
B.~Kiessling, ``Kraken-an universal text recognizer for the humanities,'' in \emph{ADHO, {\'E}d., Actes de Digital Humanities Conference}, 2019.

\bibitem{aguilar2025dual}
S.~T. Aguilar, ``Dual-style transcription of historical manuscripts based on multimodal small language models with switchable adapters,'' 2025.

\bibitem{stutzmann_dominique_2021_5535306}
\BIBentryALTinterwordspacing
D.~Stutzmann, S.~Hamel, I.~d. Kernier, G.~Mühlberger, and G.~Hackl, ``Himanis guérin,'' Sep. 2021. [Online]. Available: \url{https://doi.org/10.5281/zenodo.5535306}
\BIBentrySTDinterwordspacing

\bibitem{stutzmann2021home}
D.~Stutzmann, S.~T. Aguilar, and P.~Chaffenet, ``Home-alcar: Aligned and annotated cartularies,'' 2021.

\bibitem{claustre_2023_7575693}
\BIBentryALTinterwordspacing
J.~Claustre, D.~Smith, S.~Torres~Aguilar \emph{et~al.}, ``The e-ndp project : collaborative digital edition of the chapter registers of notre-dame of paris (1326-1504). ground-truth for handwriting text recognition (htr) on late medieval manuscripts.'' Feb. 2023. [Online]. Available: \url{https://doi.org/10.5281/zenodo.7575693}
\BIBentrySTDinterwordspacing

\bibitem{scius2023bullinger}
A.~Scius-Bertrand, P.~Str{\"o}bel, M.~Volk, T.~Hodel, and A.~Fischer, ``The bullinger dataset: A writer adaptation challenge,'' in \emph{International Conference on Document Analysis and Recognition}.\hskip 1em plus 0.5em minus 0.4em\relax Springer, 2023, pp. 397--410.

\bibitem{clerice2024catmus}
T.~Cl{\'e}rice, A.~Pinche, M.~Vlachou-Efstathiou, A.~Chagu{\'e} \emph{et~al.}, ``Catmus medieval: A multilingual large-scale cross-century dataset in latin script for handwritten text recognition and beyond,'' in \emph{International Conference on Document Analysis and Recognition}.\hskip 1em plus 0.5em minus 0.4em\relax Springer, 2024, pp. 174--194.

\bibitem{fischer2011transcription}
A.~Fischer, V.~Frinken, A.~Forn{\'e}s, and H.~Bunke, ``Transcription alignment of latin manuscripts using hidden markov models,'' in \emph{Proceedings of the 2011 Workshop on Historical Document Imaging and Processing}, 2011, pp. 29--36.

\bibitem{fischer2009automatic}
A.~Fischer, M.~Wuthrich, M.~Liwicki, V.~Frinken, H.~Bunke, G.~Viehhauser, and M.~Stolz, ``Automatic transcription of handwritten medieval documents,'' in \emph{2009 15th International Conference on Virtual Systems and Multimedia}.\hskip 1em plus 0.5em minus 0.4em\relax IEEE, 2009, pp. 137--142.

\bibitem{romero2013esposalles}
V.~Romero, A.~Forn{\'e}s, N.~Serrano, J.~A. S{\'a}nchez, A.~H. Toselli, V.~Frinken, E.~Vidal, and J.~Llad{\'o}s, ``The esposalles database: An ancient marriage license corpus for off-line handwriting recognition,'' \emph{Pattern Recognition}, vol.~46, no.~6, pp. 1658--1669, 2013.

\bibitem{driscoll}
M.~Driscoll, ``\BIBforeignlanguage{English}{Levels of transcription},'' in \emph{\BIBforeignlanguage{English}{Electronic textual editing}}.\hskip 1em plus 0.5em minus 0.4em\relax Modern Language Association of America, 2006, pp. 254--261.

\bibitem{hasenohr1998abreviations}
G.~Hasenohr, ``Abr{\'e}viations et fronti{\`e}res de mots,'' \emph{Langue fran{\c{c}}aise}, pp. 24--29, 1998.

\bibitem{borja2012desarrollo}
P.~S.-P. Borja, ``Desarrollo y explotación del corpus de documentos españoles anteriores a 1700(codea),'' \emph{Scriptum digital}, no.~1, pp. 5--35, 2012.

\bibitem{halter_pernet_2021_5179361}
\BIBentryALTinterwordspacing
C.~Halter-Pernet, S.~Teuscher, T.~Hodel, L.~Barwitzki, S.~Egloff, F.~Henggeler, M.~Nadig, A.~Steinmann, S.~Stettler, and I.~Prada~Ziegler, ``{Charters and Records of Königsfelden Abbey and Bailiwick (1308-1662)},'' Oct. 2021. [Online]. Available: \url{https://doi.org/10.5281/zenodo.5179361}
\BIBentrySTDinterwordspacing

\bibitem{https://doi.org/10.5281/zenodo.6414086}
\BIBentryALTinterwordspacing
L.~Keijser, ``\BIBforeignlanguage{odt}{6000 ground truth of voc and notarial deeds 3.000.000 htr of voc, wic and notarial deeds},'' 2020. [Online]. Available: \url{https://zenodo.org/doi/10.5281/zenodo.6414086}
\BIBentrySTDinterwordspacing

\bibitem{alkendi2024advancements}
W.~AlKendi, F.~Gechter, L.~Heyberger, and C.~Guyeux, ``Advancements and challenges in handwritten text recognition: A comprehensive survey,'' \emph{Journal of Imaging}, vol.~10, no.~1, p.~18, 2024.

\end{thebibliography}

\end{document}